\documentclass[10pt,conference]{IEEEtran}
\usepackage{amsmath, amssymb}
\usepackage{algorithm}
\usepackage{algorithmicx}
\usepackage{algcompatible}

\usepackage{amsmath, amssymb}

\usepackage{algorithm}
\usepackage{algpseudocode}

\usepackage{booktabs, makecell, multirow, siunitx, threeparttable}
\usepackage{graphicx}
\usepackage{subcaption}
\usepackage{float}
\usepackage{tikz}
\usepackage[most]{tcolorbox}
\usepackage{adjustbox}
\usepackage{pdflscape}

\usepackage{enumitem}
\usepackage{titlesec}
\titlespacing*{\subsubsection}{0pt}{\baselineskip}{\baselineskip}

\usepackage{xcolor}
\usepackage{hyperref}

\usepackage{cite}

\graphicspath{{images/}}

{\begin{list}{}%
        {\setlength{\leftmargin}{#1}}%
       \item[]%
}
{\end{list}}

\begin{document}

\title{Traceability and Accountability in Role-Specialized Multi-Agent LLM Pipelines}







\author{
\IEEEauthorblockN{Amine Barrak}
\IEEEauthorblockA{\textit{Department of Computer Science and Engineering} \\
Oakland University, Rochester, MI, USA}
\IEEEauthorblockA{aminebarrak@oakland.edu}
}

\maketitle

\newcommand{\emna}[1]{\textcolor{red}{#1}}

\begin{abstract}
Sequential multi-agent systems built with large language models (LLMs) can automate complex software tasks, but they are hard to trust because errors quietly pass from one stage to the next. We study a traceable and accountable pipeline, meaning a system with clear roles, structured handoffs, and saved records that let us trace who did what at each step and assign blame when things go wrong. Our setting is a Planner → Executor → Critic pipeline. We evaluate eight configurations of three state-of-the-art LLMs on three benchmarks and analyze where errors start, how they spread, and how they can be fixed. Our results show: (1) adding a structured, accountable handoff between agents markedly improves accuracy and prevents the failures common in simple pipelines; (2) models have clear role-specific strengths and risks (e.g., steady planning vs. high-variance critiquing), which we quantify with repair and harm rates; and (3) accuracy–cost–latency trade-offs are task-dependent, with heterogeneous pipelines often the most efficient. Overall, we provide a practical, data-driven method for designing, tracing, and debugging reliable, predictable, and accountable multi-agent systems.
\end{abstract}

\vspace{0.1cm}

\section{Introduction}

The field of software engineering is undergoing a significant transformation, driven by the rise of Large Language Models (LLMs). This evolution is progressing from LLM-powered ``co-pilots'' that assist developers to fully autonomous, multi-agent systems capable of tackling complex software development tasks with minimal human intervention~\cite{he2025llm, jin2024llms}. A common architectural pattern is the \textit{sequential multi-agent pipeline}, where specialized LLM-based agents collaborate in a predefined order to perform roles such as planning, development, and testing~\cite{wang2024survey}. Systems like ChatDev and MetaGPT have demonstrated the potential of this approach, where task decomposition allows multiple agents to synergistically solve problems that would be intractable for a single monolithic model~\cite{qian2023communicative, hong2024metagpt}.

\begin{figure}[t]
\centering
\includegraphics[width=1\linewidth]{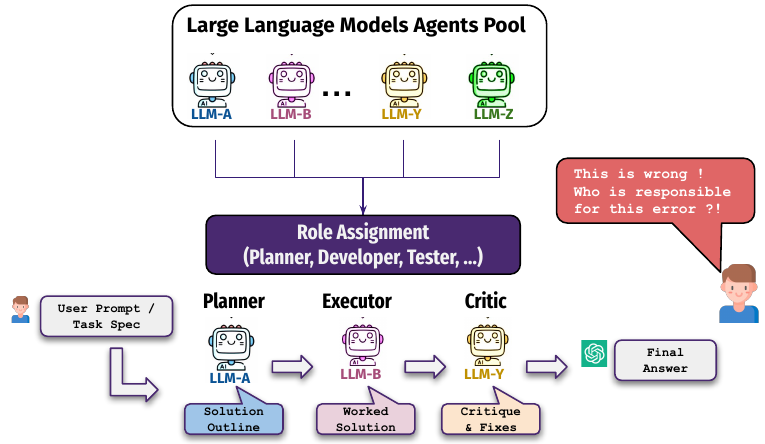}
\caption{A sequential multi-agent pipeline. A failure in the final output makes debugging and assigning responsibility a critical challenge.}
\vspace{-10pt}
\label{fig:study_method}
\end{figure}

However, this shift to sequential agent pipelines introduces a critical software engineering challenge: the loss of transparency and the immense difficulty of debugging. When agents are chained together, an error introduced by an early agent can silently cascade, corrupting the entire workflow and leading to a final output that is incorrect for reasons that are difficult to diagnose~\cite{wang2024survey}. As illustrated in Figure~\ref{fig:study_method}, this turns the collaborative pipeline into a new kind of opaque system where failures are observable, but their origins are not. With developers already spending over 50\% of their time on debugging traditional software, the introduction of these complex, non-deterministic agentic systems threatens to exacerbate this challenge significantly~\cite{lee2024unified}. Debugging a sequential pipeline of agents is therefore a crucial, yet largely unsolved, problem standing in the way of their reliable deployment~\cite{epperson2025interactive, yang2024large}.

While the potential of multi-agent systems is well-documented, the engineering principles required to make them trustworthy and maintainable are still nascent~\cite{ronanki2025facilitating}. Much of the existing research on systems like ChatDev~\cite{qian2023communicative}, MetaGPT~\cite{hong2024metagpt}, and other debugging agents~\cite{lee2024unified} focuses on demonstrating the capabilities of agent collaboration. While these studies are foundational, they often treat the pipeline as a black box, evaluating only the final output. Consequently, they do not provide a systematic analysis of the internal system dynamics or the root causes of failure. This is a critical omission, as recent work has shown that multi-agent systems can unexpectedly underperform strong, yet simpler, single-agent baselines due to coordination overhead and these very cascading errors~\cite{pan2025multiagent}. The software engineering community has long recognized traceability as a cornerstone of building robust systems~\cite{cleland2014software}, yet this principle has not been systematically applied to the internal workings of agentic pipelines. This gap has recently been framed as the challenge of ``automated failure attribution''---identifying which agent is responsible for a failure---a task that is critical for debugging but remains a labor-intensive and underexplored problem~\cite{zhang2025agent, triantafyllou2021blame}.

To address this gap, this paper presents an in-depth empirical study of an \textbf{accountable sequential pipeline} of LLM-based agents. Rather than proposing a new framework, our contribution is a rigorous analysis of the internal dynamics of such a pipeline. We employ a blame attribution methodology to monitor the correctness of a solution as it passes through a \texttt{Planner} $\rightarrow$ \texttt{Executor} $\rightarrow$ \texttt{Critic} sequence. This allows us to quantify novel, role-specific behaviors such as \textbf{repair} (when an agent corrects an error from a preceding stage) and \textbf{harm} (when an agent introduces an error to a previously correct state). Through a large-scale study involving three state-of-the-art LLMs in eight distinct pipeline configurations across three diverse benchmarks, we systematically investigate the following research questions:
\begin{itemize}
    \item \textbf{RQ1:} How do sequential multi-agent pipelines compare to the performance of monolithic LLMs?
    \item \textbf{RQ2:} How do individual LLMs perform in specialized roles, and what are the dynamics of error propagation and correction within the pipeline?
    \item \textbf{RQ3:} What are the trade-offs between accuracy, cost, and latency when designing accountable sequential agent pipelines?
\end{itemize}

We release the dataset used in this study.\footnote{\url{https://sites.google.com/view/mas-gain2025/home}}

\section{Related Work}
\label{sec:related_work}

Our work sits at the intersection of multi-agent LLM systems for software development and verifiable feedback loops for agentic workflows, addressing a key gap: the lack of a formal, traceable, and debuggable pipeline design.

\subsection{Multi-Agent Collaboration in Autonomous Software Engineering}

The inherent limitations of single, monolithic LLMs in handling complex, multi-step tasks are well-established. A prominent issue is the "lost in the middle" problem, where models exhibit degraded performance in utilizing information positioned in the center of long context windows \cite{1}. This has catalyzed a shift towards multi-agent systems, where a complex problem is decomposed and delegated to a society of specialized, collaborative agents.

Within software engineering, this paradigm has been successfully instantiated by frameworks that simulate entire development organizations. For instance, \textbf{ChatDev} \cite{qian2023communicative} constructs a virtual software company where LLM agents embody roles like CEO, programmer, and tester, following a structured waterfall process to develop software. Similarly, \textbf{MetaGPT} \cite{hong2024metagpt} employs agents as product managers and architects, mandating the generation of standardized artifacts like requirement documents and system designs before implementation begins. This structured approach is facilitated by general-purpose frameworks like \textbf{AutoGen} \cite{4}, which provides a toolkit for orchestrating these complex multi-agent conversations. These systems convincingly demonstrate the feasibility of role-based decomposition for complex software generation.

The choice of LLM for each specialized agent is a critical design decision with economic and performance implications. Stronger models deliver higher accuracy and reasoning but incur greater cost and latency, while lighter models are faster and cheaper but less effective for planning or critical analysis \cite{chen2023agentverse}. This trade-off compels careful role allocation to balance cost, speed, and quality \cite{wang2024survey}. To automate this process, routing frameworks have emerged; for example, {\textit{CARGO}} introduces a confidence-aware system that dynamically assigns prompts to the most suitable model, managing the performance–cost trade-off without manual intervention \cite{barrak2025cargo}.

However, these frameworks often model entire organizations with complex interaction patterns. In our study, we abstract away from specific organizational structures to investigate the fundamental effectiveness of a pipelined LLM workflow. We propose to isolate and analyze a core, generalizable process by splitting a task into three distinct stages—Plan, Execution, and Critic. To formally study the impact of this role-based decomposition, we evaluate its accuracy, cost, and latency across a range of complex tasks foundational to modern software engineering.

\subsection{Traceability and Correction in Agentic Pipelines}

Beyond end-to-end simulation, a parallel stream of research has focused on the feedback mechanisms within agentic workflows. The \textbf{Self-Refine} \cite{5} framework established that an LLM can iteratively improve its own output by generating self-feedback and then acting upon it. This principle has been adapted for coding in systems like \textbf{Self-Debug} \cite{6}, where an agent is prompted to first analyze and explain a bug before attempting a fix. The role of a critic is further exemplified in automated code review agents \cite{7} and in workflows inspired by Test-Driven Development.

While these approaches validate the concept of a critique-and-refine loop, a simple pipeline is not sufficient for building robust and reliable systems. A primary challenge in current agentic systems is the lack of \textit{observability} and \textit{traceability} \cite{wang2024survey, epperson2025interactive}. When a pipeline produces a final, incorrect answer, it is often difficult to perform root cause analysis and identify which stage or agent introduced the error. This "black box" nature hinders debugging and iterative improvement. Furthermore, recent work on iterative refinement for ML pipelines has shown that modifying and evaluating one component at a time leads to more stable and interpretable improvements \cite{10}. This highlights the need for modularity, where a specific LLM in the pipeline can be evaluated and replaced if it proves to be a weak link. The concept of using multiple, diverse verifiers to evaluate an output also suggests that a robust critique stage is multi-faceted and crucial for reliability \cite{11}.

Our work directly addresses this gap. We contend that a pipeline must be designed for traceability to track the evolution of decisions and artifacts at each stage. This not only facilitates debugging and the modular replacement of underperforming agents but also enables early-stage error correction. By evaluating the output of each stage, our pipeline allows a subsequent agent to identify and rectify issues before they propagate.

\section{Design Study}
\label{sec:study-design}
To evaluate the effectiveness of the traceable multi-agent pipeline, we conducted a controlled study across multiple datasets, model configurations, and prompt conditions. This section outlines the design choices and experimental setup.

\subsection{Prompts Datasets Collection}

We curated multiple-choice datasets, retaining only prompts with valid ground-truth answers to ensure consistent evaluation of accuracy across role configurations and blame attributions. All prompts were aligned to the standard answer space (\(\{A,B,C,D,E\}\)).

We evaluate on three complementary benchmarks:
\begin{itemize}
  \item \textbf{PythonIO} (127 prompts): programming and algorithmic reasoning problems expressed in natural language, stressing systematic reasoning and code-like logic \cite{zhang2024mcpythonio}.  
  \item \textbf{LogiQA} (283 prompts): formal logic and deductive reasoning tasks; originally released by Liu et al. (2020) and publicly available via GitHub \cite{liu2020logiqa}.  
  \item \textbf{AGIEval} (263 prompts): standardized test–style questions (math, reading, commonsense); derived from public exam data (e.g. Gaokao, SAT) and repurposed in AGIEval benchmark \cite{zhong2023agieval}.  
\end{itemize}

\subsection{Model Selection Rationale}
For this study, we selected three frontier LLMs that held top positions on public leaderboards during our evaluation period:
\begin{itemize}
  \item \textbf{GPT-4o (A).} Reported at the top of the Chatbot Arena (LMSYS) overall leaderboard in mid–2024 \cite{forbes_arena_2024}. \emph{Noted for stable, low-variance responses and strong text+code reasoning.}
  \item \textbf{Claude 3.5 Sonnet (B).} Public reports placed it at the top of key LMSYS Chatbot Arena categories shortly after release (June~2024) \cite{venturebeat_claude35_2024}. \emph{Noted for strict instruction-following and effective multi-step revision/correction.}
  \item \textbf{Gemini 2.5 Pro (C).} Covered as leading the LMArena (human-preference) leaderboard in early 2025 \cite{verge_gemini25_2025}. \emph{Noted for efficient high-capability reasoning and coding with reliable schema compliance in MCQ settings.}
\end{itemize}

\subsection{Experimental Conditions}
We compare three inference regimes:

\noindent\underline{\textbf{(1) Single-model baselines.}} 
Each model answers end-to-end with the prompt above. These runs establish individual capability and serve as reference points.

\medskip
\noindent\underline{\textbf{(2) Pipelines without traceability.}} 
We compose three-step pipelines (length \(3\)) using ordered model triples from \(\{A,B,C\}\), e.g.,
\(\mathrm{AAA}, \mathrm{BBB}, \mathrm{CCC}, \mathrm{ABC}, \mathrm{ACB}, \ldots\).
Each step consumes the previous step's artifact (question and current answer) and emits a new one-letter answer.
We score only the final output.

\medskip
\noindent\underline{\textbf{(3) Pipelines with traceability (\emph{blame tracking}).}} 
We instrument the same configurations to capture stagewise inputs/outputs and deltas against gold labels.
For each item we compute whether the intermediate step \emph{repaired} a wrong answer, \emph{harmed} a correct answer,
or performed a \emph{no-op}.

\subsection{Evaluation Metrics}
\label{sec:metrics}

To conduct a comprehensive evaluation of both monolithic models and the sequential agent pipelines, we compute a set of quantitative metrics designed to capture effectiveness, efficiency, and operational cost.

\noindent\underline{\textbf{Token statistics}}
We collect detailed token usage for each model interaction, including the number of input (prompt) tokens and output (completion) tokens. This data is essential for analyzing the complexity of the tasks given to each agent and serves as the basis for cost computation. We report the median to provide a robust measure of token consumption.

\noindent\underline{\textbf{Accuracy}}
This is the primary metric for evaluating the correctness of the final output. For each task, the generated answer is compared against a ground-truth solution. Accuracy is calculated as the percentage of tasks for which the model or pipeline produced the correct final answer.

\noindent\underline{\textbf{Cost Computation}}
To assess the operational expense of each approach, we compute the monetary cost. Let \(R_{\text{in}}(m)\) and \(R_{\text{out}}(m)\) be the provider’s prices (USD) per 1{,}000 tokens for input and output respectively for model \(m\). For a dataset \(D\), with total prompt tokens \(T_{\text{in}}(m,D)\) and total completion tokens \(T_{\text{out}}(m,D)\) produced by model \(m\), the exact total cost we report is:
\

\noindent
When per-call cost is needed, it is computed itemwise as
\(
\frac{\text{prompt}_i}{1000}R_{\text{in}}(m)
+ \frac{\text{completion}_i}{1000}R_{\text{out}}(m)
\)
and summarized by the median across items. For the sequential multi-agent pipeline, the total cost for a single task is the sum of the costs incurred by each agent:
\[ 
\text{Pipeline Cost} = \text{Cost}_\text{planner} + \text{Cost}_\text{executor} + \text{Cost}_\text{critic} 
\]
All reported monetary totals were computed with the fixed per-token rates detailed in Table~\ref{tab:pricing}.

\begin{table}[h]
\centering
\small
\caption{Pricing comparison of major LLMs with input and output costs per 1K tokens.}
\label{tab:pricing}
\begin{adjustbox}{width=\linewidth}
\begin{tabular}{lrrr}
\toprule
\textbf{Model} & 
\makecell{Input \\ (USD / 1K tokens)} & 
\makecell{Output \\ (USD / 1K tokens)} & 
Source \\
\midrule
GPT\texttt{-}4o (A) & \$0.0050 & \$0.0200 & \cite{gpt_price}\\
Claude Sonnet 4 (B) & \$0.0030 & \$0.0150 & \cite{claude_price}\\
Gemini 2.5 Pro (C) & \$0.00125 & \$0.0100 & \cite{gemini_price}\\
\bottomrule
\end{tabular}
\end{adjustbox}
\end{table}

\noindent\underline{\textbf{Latency statistics}}
To measure the efficiency and responsiveness of each configuration, we record the end-to-end latency for each task in seconds. This includes the time taken for all API calls and intermediate processing within the pipeline. We report the median latency to provide a stable measure of the system's response time.

\subsection{Blame Assignment in Multi-Agent Pipelines}
We use a three–role pipeline: a \textbf{Planner} proposes an initial answer, an \textbf{Executor} solves the task conditioned on the planner’s output, and a \textbf{Critic} reviews or revises the executor’s answer. The pipeline publishes a single final answer by preferring later stages (critic $\rightarrow$ executor $\rightarrow$ planner). To analyze accountability, we define a \emph{blame function} that compares each stage’s answer to the ground truth and sets binary flags indicating whether a downstream stage \emph{repaired} a wrong upstream answer or \emph{harmed} a correct one; the \emph{error origin} is the earliest stage whose mistake remains unrepaired in the final output (NONE if the final answer is correct). The operational logic is summarized below.

\begin{algorithm}[H]
\caption{Traceable Pipeline (Roles and Blame Logic)}
\label{alg:traceable-pipeline}
\begin{algorithmic}[1]
\Require Dataset $D=\{(x_i,y_i)\}_{i=1}^N$; models $M_P$ (Planner), $M_E$ (Executor), $M_C$ (Critic)
\Ensure For each $i$: stage answers $(P,E,C)$, final $F$, blame flags, error origin

\For{$i \gets 1$ \textbf{to} $N$}
  \State $P \gets M_P(x_i)$
  \State $E \gets M_E(x_i, P)$
  \State $C \gets M_C(x_i, P, E)$

  \State \textbf{Final selection}:
  \If{$C$ is defined}
    \State $F \gets C$
  \ElsIf{$E$ is defined}
    \State $F \gets E$
  \Else
    \State $F \gets P$
  \EndIf

  \State \textbf{Blame flags}:
  \State $\texttt{planner\_error}[i] \gets (P \neq y_i)$
  \State $\texttt{executor\_repair}[i] \gets (P \neq y_i) \land (E = y_i)$
  \State $\texttt{executor\_harm}[i] \gets (P = y_i) \land (E \neq y_i)$
  \State $\texttt{critic\_repair}[i] \gets (E \neq y_i) \land (C = y_i)$
  \State $\texttt{critic\_harm}[i] \gets (E = y_i) \land (C \neq y_i)$

  \State \textbf{Error origin}:
  \If{$F = y_i$}
    \State $\texttt{origin}[i] \gets \text{NONE}$
  \Else
    \If{$(E = y_i) \land (C \neq y_i)$}
      \State $\texttt{origin}[i] \gets \text{CRITIC}$
    \ElsIf{$(P = y_i) \land (E \neq y_i)$}
      \State $\texttt{origin}[i] \gets \text{EXECUTOR}$
    \Else
      \State $\texttt{origin}[i] \gets \text{PLANNER}$
    \EndIf
  \EndIf

  \State \textbf{Record} $(P,E,C,F,\text{flags},\texttt{origin}[i])$
\EndFor
\end{algorithmic}
\end{algorithm}

\section{Results}
\label{sec:results}

This section presents the empirical findings from our evaluation of monolithic and sequential multi-agent pipeline configurations across three distinct benchmarks: AgiEval (general reasoning), PythonIO (code generation), and LogiQA (logical reasoning). The results are organized to systematically address our three primary research questions.

\subsection{{RQ1:} How do sequential multi-agent pipelines compare to the performance of monolithic LLMs?}
To justify the use of multi-agent systems, we first establish a performance baseline by evaluating each LLM as a monolithic, end-to-end problem solver. We then compare these results against two types of sequential pipelines—a simple, unmonitored pipeline and an accountable pipeline with a structured handoff protocol—to determine the impact of both collaboration and accountability.

\subsubsection{Baseline Performance of Monolithic Models}
We began by assessing the individual capabilities of our three selected LLMs—Gemini 2.5 Pro (Model C), Claude Sonnet 4 (Model B), and GPT-4o (Model A)—on each benchmark. The results, shown in Table~\ref{tab:baseline}, summarizes the accuracy, median latency, and median cost per prompt for each model.


\begin{table}[h]
\centering
\small
\caption{Baseline Performance of Single Models Across All Datasets}
\begin{adjustbox}{width=\linewidth}
\begin{tabular}{@{}llccc@{}} 
\toprule 
\textbf{Dataset} & \textbf{Model} & \textbf{Accuracy (\%)} & 
\makecell{\textbf{Median}\\\textbf{Latency (s)}} & 
\makecell{\textbf{Median}\\\textbf{Cost/Prompt (USD)}} \\ 
\midrule 
\textbf{AgiEval} 
 & Gemini 2.5 Pro (C) & 92.40 & 11.84 & \$0.0112 \\ 
 & Claude Sonnet 4 (B) & 58.20 & 4.21 & \$0.0036 \\ 
 & GPT-4o (A)         & 43.70 & 0.74 & \$0.0006 \\ 
\midrule 
\textbf{PythonIO} 
 & Gemini 2.5 Pro (C) & 99.21 & 11.20 & \$0.0120 \\ 
 & Claude Sonnet 4 (B) & 53.54 & 5.48 & \$0.0054 \\ 
 & GPT-4o (A)         & 69.29 & 0.71 & \$0.0009 \\ 
\midrule 
\textbf{LogiQA} 
 & Gemini 2.5 Pro (C) & 84.45 & 14.78 & \$0.0134 \\ 
 & Claude Sonnet 4 (B) & 79.51 & 6.74 & \$0.0051 \\ 
 & GPT-4o (A)         & 67.49 & 0.77 & \$0.0009 \\ 
\bottomrule 
\end{tabular} 
\end{adjustbox}

\label{tab:baseline}
\end{table}

The results in Table~\ref{tab:baseline} establish a clear performance hierarchy, with Gemini 2.5 Pro representing the best-case scenario for a single-model, monolithic approach. While its accuracy is formidable, the increasing complexity of agentic AI tasks is beginning to challenge the limits of single-agent systems, particularly for long-horizon problems that require intricate planning and reasoning \cite{wang2024chainofagents, mushtaq2025harnessing}. Fields such as autonomous software development and multi-step scientific discovery present significant hurdles for monolithic models \cite{he2025llm}. For such intricate problems, a monolithic model faces significant hurdles: it can struggle to maintain context across numerous steps, its reasoning process for intermediate sub-problems remains opaque, and a single early error can irrevocably corrupt the entire solution without any mechanism for verification or recovery. This paradigm of growing task complexity motivates our investigation into a decomposed, multi-agent pipeline. The central hypothesis is not to compete with the raw accuracy of the best single model on constrained benchmarks, but to propose a \textbf{modular and verifiable architecture} suited for complex, multi-step processes. By assigning specialized roles (e.g., planning, execution, critique) to different agents, we can construct systems designed for this purpose.

\subsubsection{Impact of Accountable Protocol on Pipeline Performance}

Having established the monolithic baselines, we next investigated the impact of the pipeline structure. We compared two architectures: a \textbf{Simple Pipeline}, where the output of one agent is passed to the next, and an \textbf{Accountable Pipeline}, which implements a structured handoff protocol with blame attribution. This protocol ensures that each agent's output is validated and passed in a consistent format, giving subsequent agents a clear state to evaluate and act upon.

Table~\ref{tab:impact} presents a direct comparison of the final accuracy of these two pipeline types across all configurations.

\begin{table}[h!]
\centering
\small
\caption{Comparative Accuracy of Simple vs.\ Accountable Pipelines Across Datasets (\%) — rows by configuration.}
\label{tab:impact}
\begin{adjustbox}{width=0.92\linewidth}
\begin{tabular}{@{}l l c c c@{}}
\toprule
\textbf{Dataset} & \textbf{Config} & \textbf{Simple Pipeline} & \textbf{Accountable Pipeline} & \textbf{Accuracy $\Delta$} \\
\midrule
\multicolumn{5}{@{}l}{\textbf{AgiEval}}\\
\cmidrule(lr){1-5}
 & AAA & 57.41 & 77.19 & +19.78 \\
 & ABC & 81.37 & 90.87 & +9.50 \\
 & BAB & 88.97 & 86.69 & $-2.28$ \\
 & BBB & 88.97 & 89.73 & +0.76 \\
 & BCC & 91.25 & 90.49 & $-0.76$ \\
 & CBA & 87.07 & 92.40 & +5.33 \\
 & CCA & 91.25 & 92.40 & +1.15 \\
 & CCC & 91.63 & 93.54 & +1.91 \\
\midrule
\multicolumn{5}{@{}l}{\textbf{PythonIO}}\\
\cmidrule(lr){1-5}
 & AAA & 73.23 & 88.98 & +15.75 \\
 & ABC & 99.21 & 99.21 & 0.00 \\
 & BAB & 63.78 & 96.85 & +33.07 \\
 & BBB & 61.42 & 97.64 & +36.22 \\
 & BCC & 99.21 & 98.43 & $-0.78$ \\
 & CBA & 71.65 & 99.21 & +27.56 \\
 & CCA & 71.65 & 98.43 & +26.78 \\
 & CCC & 99.21 & 99.21 & 0.00 \\
\midrule
\multicolumn{5}{@{}l}{\textbf{LogiQA}}\\
\cmidrule(lr){1-5}
 & AAA & 71.48 & 71.38 & $-0.10$ \\
 & ABC & 79.58 & 80.57 & +0.99 \\
 & BAB & 78.52 & 79.86 & +1.34 \\
 & BBB & 79.58 & 79.51 & $-0.07$ \\
 & BCC & 82.75 & 83.39 & +0.64 \\
 & CBA & 81.34 & 83.75 & +2.41 \\
 & CCA & 78.52 & 82.28 & +3.76 \\
 & CCC & 81.34 & 85.16 & +3.82 \\
\bottomrule
\end{tabular}
\end{adjustbox}
\end{table}

The data in Table~\ref{tab:impact} shows that the introduction of the accountable protocol leads to notable changes in pipeline accuracy, with the magnitude and direction of the change varying by configuration and dataset.

On the AgiEval benchmark, the largest accuracy increase was observed in the AAA configuration, which rose by 19.78 percentage points from 57.41\% to 77.19\%. The ABC configuration also saw a significant gain of 9.50 points. Most other configurations showed smaller positive changes, such as CCC (+1.91 points) and CBA (+5.33 points). Conversely, the BAB and BCC configurations saw minor decreases of 2.28 and 0.76 points, respectively.

On the PythonIO benchmark, the accuracy gains were even more substantial for several configurations. The BBB configuration's accuracy increased by 36.22 percentage points, from 61.42\% to 97.64\%. Similarly large gains were recorded for BAB (+33.07 points), CBA (+27.56 points), and CCA (+26.78 points). The top-performing configurations, ABC and CCC, showed no change, as both the simple and accountable versions achieved 99.21\% accuracy. Only one configuration, BCC, showed a small decrease of 0.78 points.

For the LogiQA benchmark, the changes were generally positive but more modest. The largest improvement was seen in the CCA configuration, which increased by 6.99 percentage points. Other configurations like CCC and CBA saw smaller gains of 3.82 and 2.41 points, respectively. The AAA and BBB configurations showed negligible negative changes of -0.10 and -0.07 points.

\begin{figure}[h]
  \centering
  \includegraphics[width=0.92\linewidth]{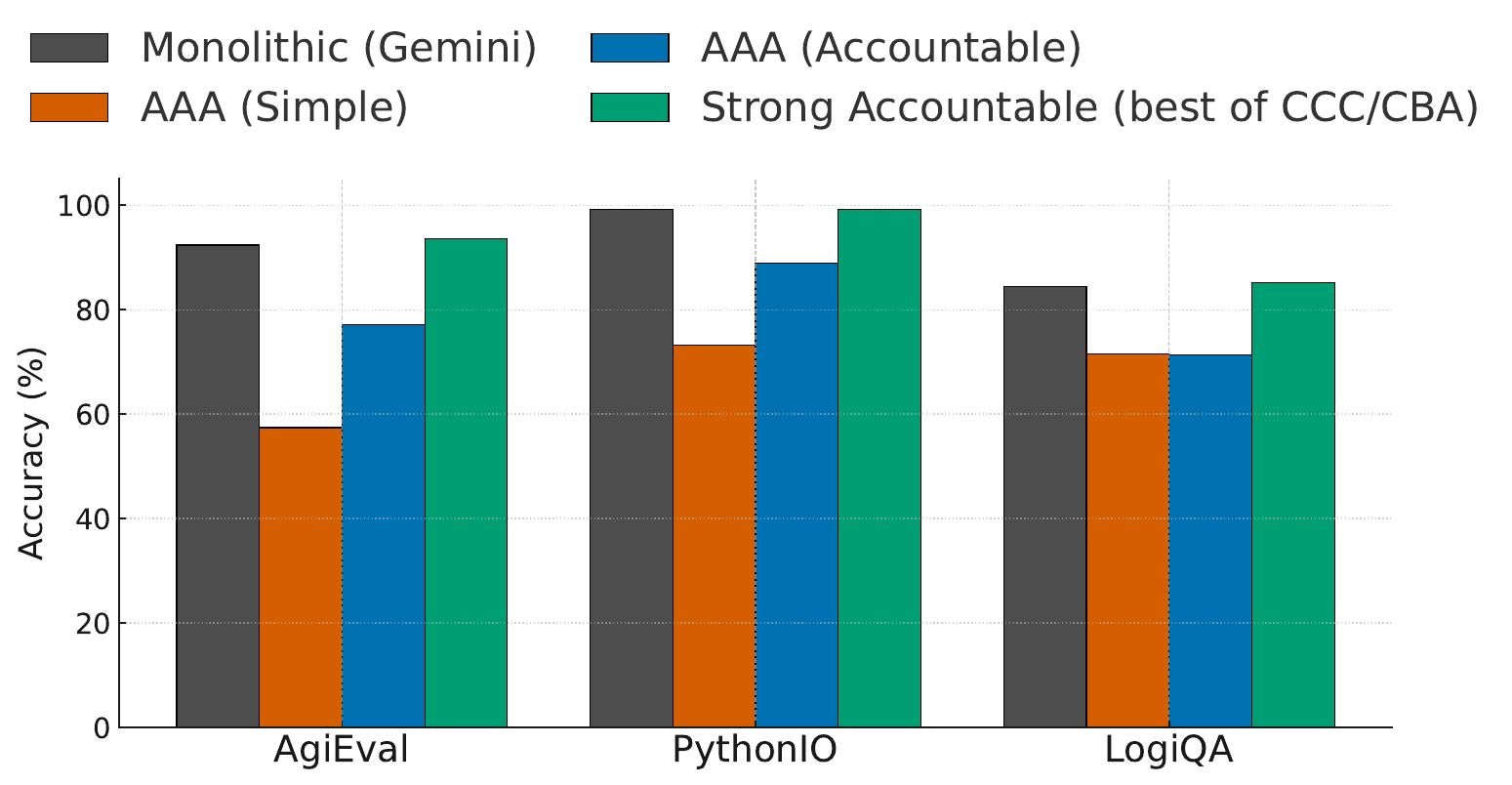}
\caption{Impact of pipeline architecture on accuracy.}
  \label{fig:impact-pipeline-accuracy}
\end{figure}

Figure \ref{fig:impact-pipeline-accuracy} provides a visual summary of these key performance trade-offs, comparing the monolithic baseline against the weakest and strongest pipeline configurations. The chart clearly illustrates three core findings across the datasets. First, the AAA (Simple) pipeline consistently underperforms, demonstrating the significant risk of "anti-synergy" where unstructured collaboration is actively harmful. Second, the AAA (Accountable) pipeline shows a dramatic performance improvement over its simple counterpart on AgiEval and PythonIO, highlighting the effectiveness of the accountable protocol in raising the performance floor. Finally, the Strong Accountable pipeline (representing the best of the CCC and CBA configurations) consistently performs at or near the level of the best monolithic model, and in the case of LogiQA, slightly surpasses it.

\begin{tcolorbox}[
  title={Findings for RQ1},
  colback=gray!5,
  colframe=black,
  boxrule=0.4pt,
  arc=2mm,
  left=2mm,right=2mm,top=1mm,bottom=1mm,
  enhanced, breakable
]
\textbf{Unstructured pipelines degrade performance.} Without clear roles and checks, errors compound across stages and accuracy falls below a competent single model.

\textbf{Accountability lifts the floor and ceiling.} A structured, accountable handoff reliably improves accuracy and stability, with strong accountable pipelines matching or surpassing weaker monolithic baselines.
\end{tcolorbox}

\subsection{RQ2: How do individual LLMs perform in specialized roles, and what are the dynamics of error propagation and correction within the pipeline?}
To address \textbf{RQ2}, we analyzed the internal dynamics of error origination and correction. 

\subsubsection{Error Origination: The Primacy of the Planner}

Our analysis reveals that the vast majority of unrecoverable pipeline failures originate from a flawed plan created in the very first step. To quantify this, we measured the error rate for each model when it was assigned the Planner role. The "Total Cases" in Table \ref{tab:planner-error-rate} represents the total number of times each model acted as the Planner across all relevant configurations for a given dataset. For example, since GPT-4o (Model A) was the Planner in two configurations (AAA and ABC), its total cases for the AgiEval benchmark (which has 263 instances) is 2 * 263 = 526.

\begin{table}[h]
\centering
\caption{Planner Error Rate by Model and Dataset.}
\label{tab:planner-error-rate}
\small
\begin{adjustbox}{width=\linewidth}
\begin{tabular}{@{}llccc@{}}
\toprule
\textbf{Planner Model} & \textbf{Dataset} & \textbf{Total Cases} & \makecell{\textbf{Errors} \\ \textbf{Introduced}} & \makecell{\textbf{Error Rate} \\ \textbf{(\%)}} \\
\midrule
\multirow{3}{*}{Gemini 2.5 Pro (C)}
& AgiEval  & 789 &  58 &  7.35 \\
& PythonIO & 381 &   3 &  0.79 \\
& LogiQA   & 849 & 129 & 15.19 \\
\midrule
\multirow{3}{*}{Claude Sonnet 4 (B)}
& AgiEval  & 789 & 106 & 13.43 \\
& PythonIO & 381 &  15 &  3.94 \\
& LogiQA   & 849 & 187 & 22.03 \\
\midrule
\multirow{3}{*}{GPT\texttt{-}4o (A)}
& AgiEval  & 526 & 213 & 40.49 \\
& PythonIO & 254 &  28 & 11.02 \\
& LogiQA   & 566 & 162 & 28.62 \\
\bottomrule
\end{tabular}
\end{adjustbox}
\end{table}

The results in Table 3 show a clear and consistent performance hierarchy among the models in the Planner role: Gemini is the most reliable, followed by Claude, with GPT-4o being the least reliable. This trend holds across all three datasets.

The error rates are also highly task-dependent. On the structured PythonIO coding task, all models performed exceptionally well as planners. Gemini's error rate was less than 1\%, and even the weakest planner, GPT-4o, only introduced an error in 11.02\% of cases. This suggests that for well-defined tasks like code generation, the planning stage is less prone to failure.

In contrast, performance degraded significantly on the more abstract reasoning tasks. On LogiQA, all models struggled more, with Claude's error rate climbing to 22.03\% and GPT-4o's to 28.62\%. The most challenging benchmark for planners was AgiEval, where GPT-4o's error rate reached a substantial 40.49\%.

\subsubsection{Mid-stream Dynamics: Quantifying Repair and Harm}
While the Planner sets the stage, the Executor and Critic roles determine the pipeline's resilience. Our blame attribution methodology allows us to measure two key behaviors: the repair rate (the frequency with which an agent corrects an error from the preceding stage) and the harm rate (the frequency with which an agent incorrectly modifies a correct output from the preceding stage). Table \ref{tab:role-repair-harm} summarizes these behavioral metrics, aggregated across all experiments.

\begin{figure*}[t!]
  \centering
  \begin{minipage}[t]{0.32\textwidth}
    \centering
    \includegraphics[width=\linewidth]{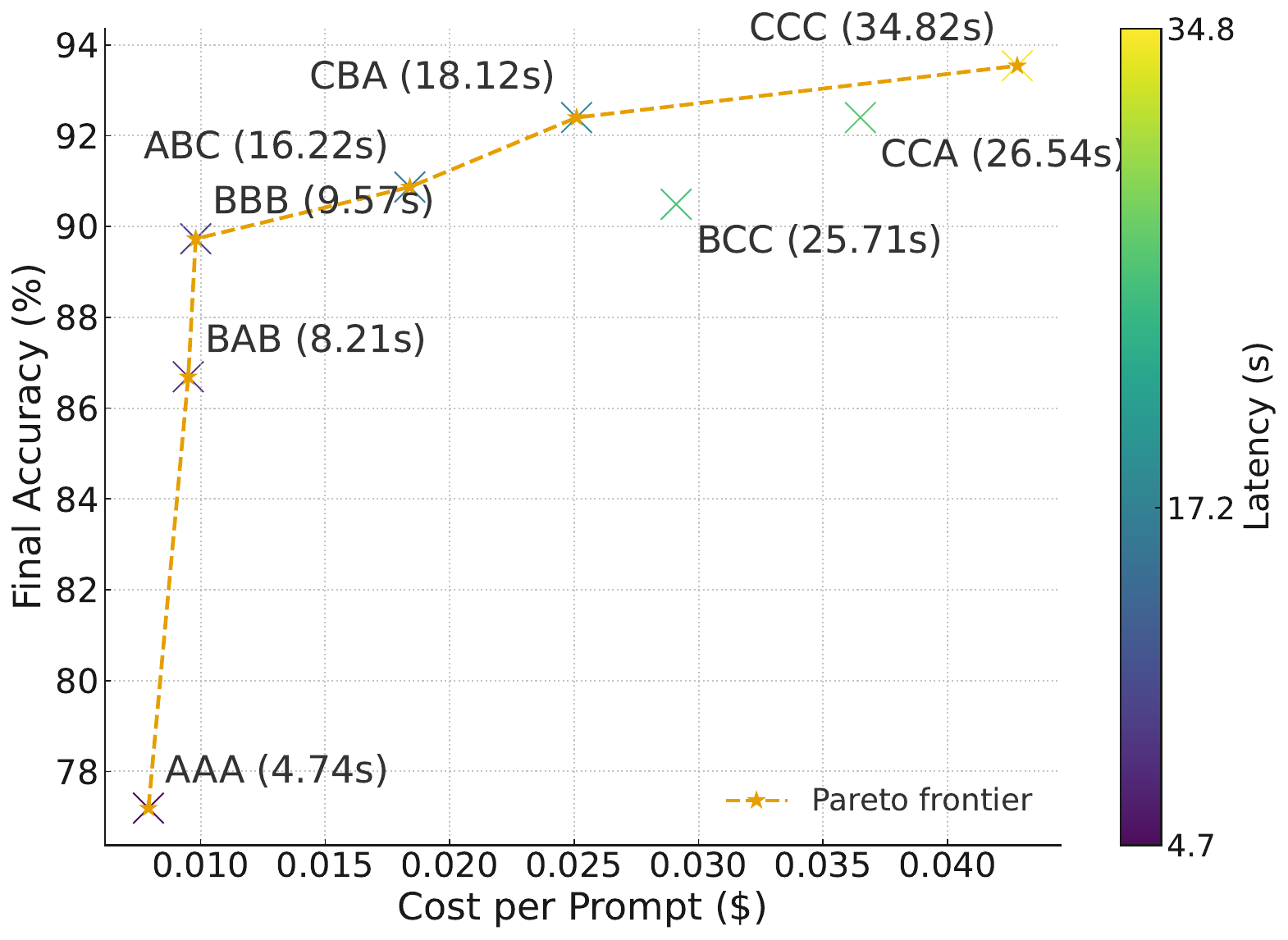}\\
    \vspace{2pt}\small AgiEval
  \end{minipage}\hfill
  \begin{minipage}[t]{0.32\textwidth}
    \centering
    \includegraphics[width=\linewidth]{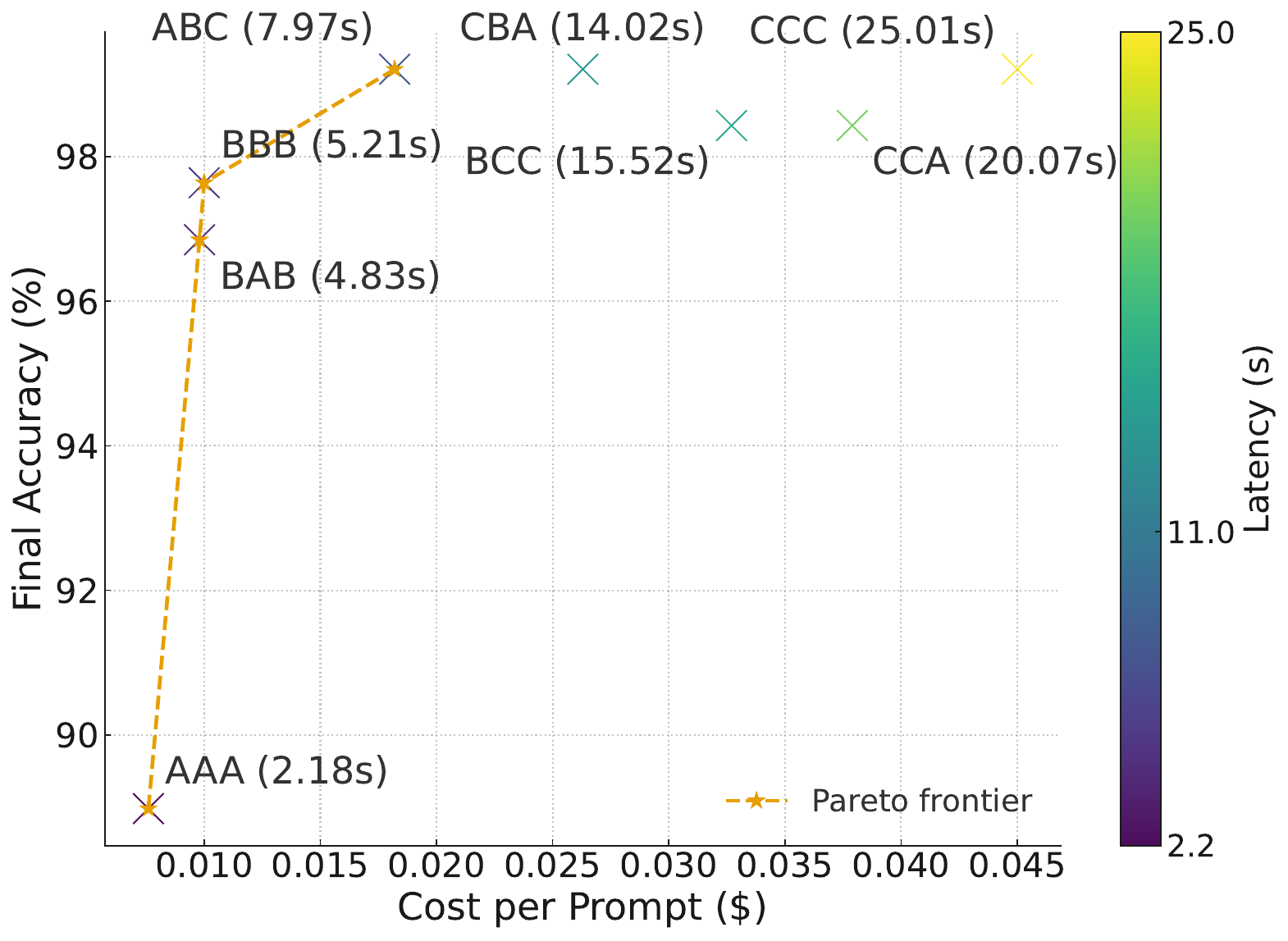}\\
    \vspace{2pt}\small PythonIO
  \end{minipage}\hfill
  \begin{minipage}[t]{0.32\textwidth}
    \centering
    \includegraphics[width=\linewidth]{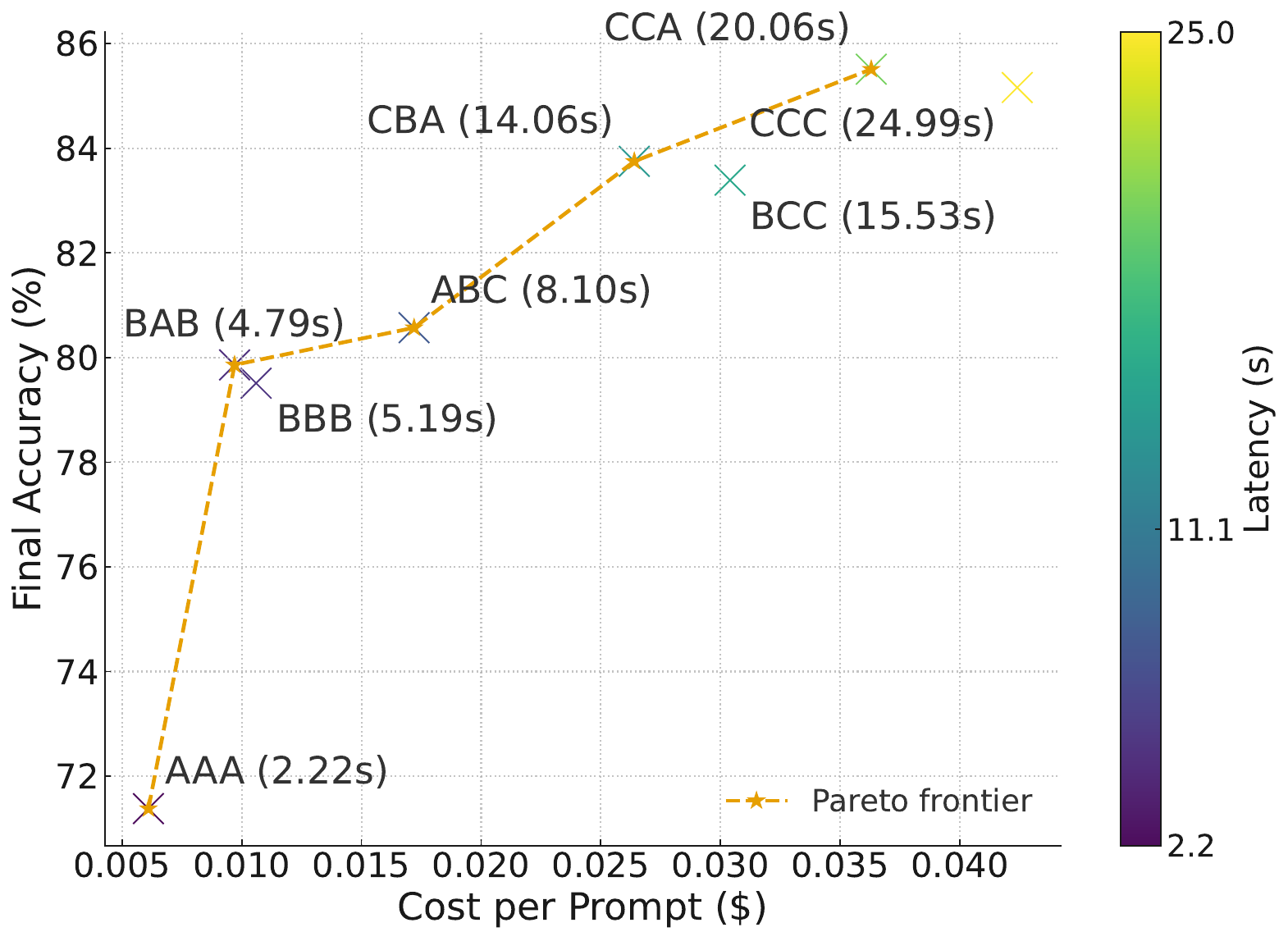}\\
    \vspace{2pt}\small LogiQA
  \end{minipage}
  \caption{Accuracy vs. cost with latency color-coded; Pareto frontier shown (dashed).}
  \label{fig:tradeoff-row-minipage}
\end{figure*}

\begin{table}[h]
\centering
\caption{Role-Specific Repair and Harm Rates by Model (Aggregated).}
\label{tab:role-repair-harm}
\small
\begin{adjustbox}{width=\linewidth}
\begin{tabular}{@{}lllcc@{}}
\toprule
\textbf{Model} & \textbf{Role} & &
\makecell{\textbf{Repair Rate}\\\textbf{(\%)}} &
\makecell{\textbf{Harm Rate}\\\textbf{(\%)}} \\
\midrule
\multirow{2}{*}{Gemini 2.5 Pro (C)}
  & Executor & & 1.27 & 0.25 \\
  & Critic   & & 2.66 & 0.25 \\
\midrule
\multirow{2}{*}{Claude Sonnet 4 (B)}
  & Executor & & 10.01 & 0.25 \\
  & Critic   & &  3.04 & 1.90 \\
\midrule
\multirow{2}{*}{GPT\texttt{-}4o (A)}
  & Executor & & 2.28 & 1.33 \\
  & Critic   & & 5.20 & 0.89 \\
\bottomrule
\end{tabular}
\end{adjustbox}
\end{table}

The data in Table \ref{tab:role-repair-harm} reveals distinct behavioral profiles for each model in the mid-stream roles.

Gemini 2.5 Pro (Model C) exhibited the lowest harm rates across both roles, at just 0.25\%. Its repair rates were also the lowest for an Executor (1.27\%) and modest for a Critic (2.66\%). 

Claude Sonnet 4 (Model B) was the most effective Executor, achieving a repair rate of 10.01\%, which was significantly higher than any other model in that role. It maintained a low harm rate of 0.25\% as an Executor. In the Critic role, its repair rate was 3.04\%, while its harm rate was the highest recorded for that role at 1.90\%.

GPT-4o (Model A) showed the highest repair rate in the Critic role, at 5.20\%. As an Executor, its repair rate was 2.28\%, but it had the highest harm rate in that role at 1.33\%.

\begin{tcolorbox}[
  title={Findings for RQ2},
  colback=gray!5,
  colframe=black,
  boxrule=0.4pt,
  arc=2mm,
  left=2mm,right=2mm,top=1mm,bottom=1mm,
  enhanced, breakable
]
\textbf{Planner primacy.} The Planner has the highest leverage; its error rate is the strongest predictor of pipeline failure.

\textbf{Role-specific aptitudes.} Models show distinct strengths. Gemini is a reliable generator but weak corrector; Claude is an excellent Executor; GPT\texttt{-}4o is a high-variance, high-reward Critic.

\textbf{Data-driven casting.} Cast roles by strengths, \textbf{Gemini} (Planner) $\rightarrow$ \textbf{Claude} (Executor) $\rightarrow$ \textbf{GPT\texttt{-}4o} (Critic).

\end{tcolorbox}

\subsection{RQ3: What are the trade-offs between accuracy, cost, and latency when designing accountable sequential agent pipelines?}
While maximizing accuracy is a primary goal, real-world deployments are constrained by operational budgets and response time requirements. Figure \ref{fig:tradeoff-row-minipage} visualizes the multi-dimensional trade-off space for our accountable pipeline configurations across the three benchmarks, plotting accuracy against cost, with latency represented by the color of each point. The dashed line in each plot indicates the Pareto frontier, representing the set of configurations that are optimal because no other configuration offers higher accuracy at a lower or equal cost.



The plots in Figure \ref{fig:tradeoff-row-minipage} reveal that the optimal configuration is highly dependent on the specific task. On AgiEval, there is a clear trade-off between accuracy and cost. The CCC configuration achieves the highest accuracy (93.54\%) but is also the most expensive. Configurations like CBA and CCA sit on the Pareto frontier, offering near-peak accuracy (92.40\%) at a significantly lower cost, making them excellent balanced choices. The fastest and cheapest configurations, such as AAA and BAB, offer lower accuracy but may be suitable for applications where speed and cost are the primary constraints.

On PythonIO, the trade-off dynamics are different. Several configurations (ABC, CBA, CCC) achieve near-perfect accuracy of 99.21\%. As the plot shows, these top-tier configurations are spread out horizontally, indicating that the key decision is no longer about gaining accuracy but about minimizing cost and latency. The ABC configuration is the most efficient of this group, while the CCC configuration is by far the most expensive for no additional accuracy gain, placing it well inside the Pareto frontier.

On LogiQA, the Pareto frontier is again clearly visible, showing a direct relationship between cost and accuracy. The CCA configuration provides the highest accuracy at 85.51\%. The CBA configuration offers a slightly lower accuracy (83.75\%) but is significantly cheaper, making it a strong alternative on the frontier. As with the other datasets, the AAA configuration is the most economical but also the least accurate.

Across all three datasets, the plots illustrate that there is no single "best" configuration. The choice depends on the specific requirements of the application, balancing the need for accuracy with the constraints of cost and latency. Furthermore, the analysis highlights the value of heterogeneous pipelines, as configurations like CBA and CCA frequently represent optimal points on the Pareto frontier.

\begin{tcolorbox}[
  title={Findings for RQ3},
  colback=gray!5,
  colframe=black,
  boxrule=0.4pt,
  arc=2mm,
  left=2mm,right=2mm,top=1mm,bottom=1mm,
  enhanced, breakable
]
\textbf{Trade-offs are task-dependent.} The optimal balance of accuracy, cost, and latency varies by task, reflected in the differing Pareto frontiers across datasets.\\[2pt]
\textbf{Heterogeneous pipelines offer robustness.} A mixed configuration such as \texttt{CBA} delivers strong, balanced performance and often lies on the Pareto frontier.\\[2pt]
\textbf{Accountability has a price.} The accountable protocol adds traceability and stability but can increase operational cost.
\end{tcolorbox}

\section{Discussion and Threats to Validity}

Our results show that accountability improves stability and interpretability, but it comes with trade-offs. In some cases, strict handoffs limit the flexibility of later agents, leading to small accuracy decreases. Structured outputs and logging also add overhead: compared to simple monolithic baselines, accountable pipelines increased costs by about 2--3$\times$ and raised median latency from $\sim$2s in lightweight settings to 20--25s in heavier ones (roughly 8--10$\times$ slower). Model choice further shapes this balance: stronger models yield higher accuracy but incur higher cost and more latency, while lighter models improve efficiency at the risk of reduced reasoning depth. Designing reliable pipelines therefore requires tuning role assignments to the task and deployment budget.

These findings should be interpreted in light of three main validity concerns. First, \textbf{construct validity:} we rely on multiple-choice benchmarks (PythonIO, LogiQA, AGIEval), which simplify open-ended tasks into discrete answers. This makes blame attribution feasible but may not reflect the full complexity of real-world software engineering. Second, \textbf{internal validity:} blame assignment assumes access to ground-truth labels, not always available at runtime; practical deployments would require proxies such as self-consistency checks, verifier agents, or unit tests. Finally, \textbf{external validity:} results are based on three proprietary LLMs during a fixed evaluation window. Model updates may alter performance, though we release prompts, logs, and evaluation dates to support replication and encourage extension to open-source models.

\section{Conclusion}
\label{sec:conclusion}
We present a large-scale study that opens the black box of sequential multi-agent LLM pipelines. Across three benchmarks we compare monolithic models, simple pipelines, and accountable pipelines. Key insights: (1) accountability is essential; structured handoffs prevent failure cascades and can raise accuracy by more than 36 points (PythonIO), (2) role specialization is real; planner quality dominates (best planner 7.35\% error vs worst 40.49\%) and models differ by role (Executor repair 10.01\%, Critic repair 5.20\%), which supports data driven casting, and (3) trade-offs among accuracy, cost, and latency are task dependent, with heterogeneous pipelines often on the Pareto frontier. We move from black box evaluation to a glass box engineering approach, with metrics and methods that help diagnose, debug, and optimize multi-agent systems for robust, predictable performance.

\section{Acknowledgement}\label{sec:ack}
This work was conducted in collaboration with \textbf{zuvu.ai}, an industrial partner developing AI-powered tools.

\bibliographystyle{IEEEtran}
\bibliography{references}
\end{document}